%% file: gaugerope.tex
\documentclass{article} 
\usepackage{preprint,times}

\input{math_commands.tex}

\usepackage[hidelinks]{hyperref}
\usepackage{url}
\usepackage{amsmath,amssymb,amsthm}
\usepackage{booktabs}
\usepackage{multirow}
\usepackage{graphicx}
\usepackage{xcolor}
\usepackage{float}
\usepackage{placeins}
\usepackage{etoolbox}
\definecolor{revisionBlue}{RGB}{0,64,180}
\newif\ifshowrevisions
\showrevisionsfalse
\newcommand{\rev}[1]{\ifshowrevisions\textcolor{revisionBlue}{#1}\else#1\fi}
\newenvironment{revision}{\begingroup\ifshowrevisions\color{revisionBlue}\fi}{\endgroup}
\AtBeginEnvironment{figure}{\color{black}}
\AtBeginEnvironment{figure*}{\color{black}}
\AtBeginEnvironment{table}{\color{black}}
\AtBeginEnvironment{table*}{\color{black}}
\usepackage{tikz}
\usepackage{pgfplots}
\usepackage{wrapfig}
\usetikzlibrary{arrows.meta,positioning,fit,calc,backgrounds}
\pgfplotsset{compat=1.18}
\definecolor{plotNavy}{RGB}{44,118,72}
\definecolor{plotBlue}{RGB}{78,153,96}
\definecolor{plotSky}{RGB}{154,205,158}
\definecolor{plotPale}{RGB}{225,242,224}
\definecolor{paperBlue}{RGB}{76,114,176}
\definecolor{paperOrange}{RGB}{221,132,82}
\usepackage{colortbl}
\definecolor{gaugeRow}{RGB}{235,244,253}

\usepackage{caption} 
\newtheorem{theorem}{Theorem}
\newtheorem{proposition}{Proposition}
\newtheorem{corollary}{Corollary}

\newcommand{\HomRoPE}{\mathrm{Hom}_{\mathrm{RoPE}}}

\title{GaugeRoPE: RoPE-Compatible Executable \\Latent Attention}

\author{{\small\textbf{Yuanhang Gao}\enspace\textbf{Xiangrui Yang}\enspace\textbf{Lingyuan Meng}\enspace\textbf{Yuanfeng Chen}\enspace\textbf{Dongsheng Li}}\\[4pt]
{\small National University of Defense Technology}\\[4pt]
{\small\texttt{\{gaoyh, yangxiangrui11, chenyuanfeng, dsli\}@nudt.edu.cn}}\\
{\small\texttt{mly\_edu@163.com}}}
\date{}

\begin{document}
	
	\maketitle
\fancyhead[L]{\small GaugeRoPE: RoPE-Compatible Executable Latent Attention}
	
	\begin{abstract}
		Latent KV cache compression reduces memory footprint by mapping historical keys and values into compact representations. However, generic compressed key states require reconstruction to native dimension before applying RoPE and computing attention scores, adding sequence-scaling overhead to the decode path. We characterize when a latent KV representation can serve directly as the execution state of pretrained RoPE. For RoPE-based transformers, exact static linear execution of the rotary key path requires the latent-to-native map to preserve the pretrained RoPE action through an intertwining relation, restricting exact mixing to coordinates that share the same RoPE frequency. Guided by this characterization, we build \textsc{GaugeRoPE}, a training-free executable latent KV representation. It combines a RoPE-compatible carrier that preserves the rotary action with an approximate position-independent residual key--value (RKV) correction for complementary key and value information, both evaluated without per-token dense historical K/V reconstruction. \rev{On Mistral-7B, GaugeRoPE achieves $5.85\times$ KV compression while maintaining quality close to the full-cache baseline.} Additional MHA results validate the construction beyond GQA. \rev{GaugeRoPE achieves up to $2.48\times$ the decode throughput of the vanilla uncompressed baseline through direct latent execution.}
	\end{abstract}

	\section{Introduction}
	\label{sec:intro}
	
	Autoregressive LLM~\citep{transformer} decoding maintains a growing history of keys and values that must be stored and repeatedly accessed at every generation step. At long context lengths and larger batch sizes, this KV cache becomes a substantial source of GPU memory footprint and memory traffic. Architectural designs such as GQA (Grouped-query attention)~\citep{gqa_ainslie} and MLA (Multi-head latent attention)~\citep{deepseekv2} reduce KV state by changing the attention architecture itself. Post-training methods instead operate on an already pretrained cache through KV quantization~\citep{kivi,kvquant}, token eviction~\citep{h2o,streamingllm}, or low-rank compression.
	
	Existing low-rank KV compression methods primarily optimize storage: they encode historical K/V into compact latent states, while generic compressed keys must still be reconstructed into native space for RoPE-aware scoring. Palu~\citep{palu} exploits redundancy in pretrained KV states, but its RoPE-aware key path reconstructs historical keys before native positional scoring. MLA~\citep{deepseekv2} avoids full native-key reconstruction by learning a compressed content latent together with a separate RoPE branch during pretraining. \rev{TransMLA~\citep{transmla} extends this decoupled design to pretrained GQA models, enabling MLA-style absorbed execution with a separate RoPE-bearing key branch.} For pretrained GQA and MHA (Multi-head attention) models, the underlying issue is structural: unless a latent key space is closed under the pretrained rotary action, RoPE cannot be applied directly within that space. Each decode step revisits the full history, so native-key reconstruction adds overhead that scales with sequence length and erodes the benefit of compressed storage. This motivates the following question: \rev{\emph{which static linear maps allow compressed states to execute the pretrained rotary attention path directly?}} We consider a training-free setting in which the pretrained RoPE action remains fixed and the compressed historical states are directly executable without per-token key reconstruction.
	
	Let $\mathbf{U}$ denote the static latent-to-native key map and let $\mathbf{R}_p$ 
	denote the pretrained positional action at position $p$. A generic
	low-rank decoder may mix coordinates rotated at different RoPE frequencies, so
	applying RoPE after decoding is generally not representable by a fixed latent
	transformation. \rev{Exact direct execution with original-frequency latent rotations therefore requires}
	\begin{equation}
		\mathbf{R}_p\mathbf{U} = \mathbf{U}\mathbf{R}'_p
		\qquad\text{for all positions }p,
		\label{eq:intro-intertwine}
	\end{equation}
	for some latent rotation $\mathbf{R}'_p$.
	
	We characterize the static linear latent maps that remain exactly compatible with pretrained RoPE under mild non-degeneracy assumptions. The resulting structure is frequency block-diagonal: distinct RoPE frequencies cannot mix exactly, whereas same-frequency copies across heads admit arbitrary complex-linear mixing. GQA and MHA provide such same-frequency copies across KV heads or attention heads, enabling compression within each frequency block while preserving the pretrained positional action. The same compatibility condition is necessary and sufficient for exact latent inner-product scoring in our execution model, allowing historical keys to be scored directly in latent space without materializing dense native-space keys. \rev{TransMLA’s RoRoPE introduces same-frequency real mixing, combined with balanced joint KV compression. Its real-coefficient carrier is a special case of the complex-linear family characterized here.}
	
	\begin{wrapfigure}{r}{0.4\linewidth}
		\centering
		\includegraphics[width=0.98\linewidth]{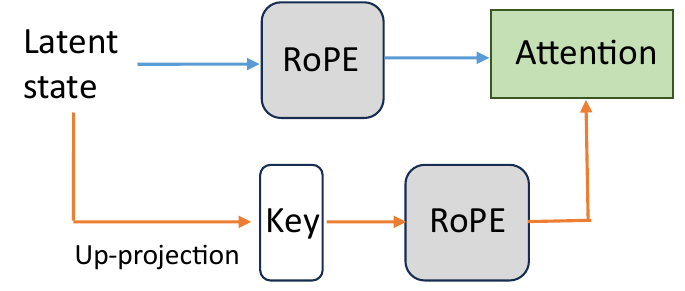}
		\caption{
			\textbf{Why RoPE compatibility matters.}
			Generic latent keys require native-space reconstruction before RoPE (orange),
			whereas GaugeRoPE supports direct latent execution (blue).
		}
		\label{fig:motivation}
	\end{wrapfigure}
	
	Figure~\ref{fig:motivation} illustrates this distinction. GaugeRoPE combines a RoPE-compatible carrier for rotary scoring with a position-independent RKV correction for residual key/value content. Both operate directly on cached latent states, avoiding per-token reconstruction of native-dimensional K/V.
	
	Experiments on GQA and MHA backbones evaluate the compatible construction and its practical benefits. \rev{At fixed rank and cache size, the additional coefficient freedom improves spectral capture, attention fidelity, and generation stability.} \rev{On Mistral, GaugeRoPE combines low-rank representation with mixed-precision quantization to achieve $5.85\times$ KV compression while maintaining quality close to the full-cache baseline.} \rev{Direct latent execution gives GaugeRoPE up to $2.48\times$ the decode throughput of the vanilla uncompressed baseline.} 
	
	\paragraph{Contributions.}
	\begin{itemize}
		\item \textbf{RoPE-compatible latent characterization.}
		We characterize the static original-frequency linear maps that carry
		pretrained RoPE exactly, showing that distinct frequencies cannot mix while
		same-frequency copies admit arbitrary complex-linear mixing.
		
		\item \textbf{Executable latent attention.}
		Guided by this characterization, GaugeRoPE combines a RoPE-compatible carrier
		with a position-independent RKV correction, enabling historical attention to
		be evaluated directly from compressed latent states without per-token native-key
		reconstruction.
		
		\item \textbf{Mechanistic and systems evaluation.}
		Experiments disentangle the roles of compatibility, approximation capacity,
		and residual correction in determining quality. System measurements
		quantify the efficiency benefits of direct latent execution.
	\end{itemize}

	\section{Problem Setup}
	\label{sec:background}
	
	\subsection{RoPE as a Positional Representation}
	\label{sec:bg-rope}
	
	RoPE~\citep{roformer} encodes position by rotating each two-dimensional query/key coordinate
	pair with a frequency-dependent phase. For one attention head with even head
	dimension $d_h$, let $F=d_h/2$ and let $\omega_j$ denote the $j$th RoPE
	frequency. Position $p$ acts on the corresponding two-dimensional plane as
	\begin{equation}
		\mathbf{R}_j(p)=
		\begin{bmatrix}
			\cos(\omega_j p) & -\sin(\omega_j p)\\
			\sin(\omega_j p) & \phantom{-}\cos(\omega_j p)
		\end{bmatrix},
		\qquad
		\mathbf{R}_p=\operatorname{diag}\bigl(\mathbf{R}_1(p),\ldots,\mathbf{R}_F(p)\bigr).
		\label{eq:rope-real}
	\end{equation}
	\rev{Here $j$ indexes frequency and $p$ position: $\mathbf{R}_j(p)$ rotates one coordinate pair, while $\mathbf{R}_p$ denotes the full positional operator.}
	For a query at position $t$ and a key at position $s$,
	\begin{equation}
		\ell_{t,s}
		=(\mathbf{R}_t\mathbf{q}_t)^\top(\mathbf{R}_s\mathbf{k}_s)
		=\mathbf{q}_t^\top \mathbf{R}_{s-t}\mathbf{k}_s,
		\label{eq:rope-relative}
	\end{equation}
	using orthogonality and $\mathbf{R}_t^\top \mathbf{R}_s=\mathbf{R}_{s-t}$.
	
	Identifying each two-dimensional plane with one complex coordinate makes the
	representation structure explicit. Write
	$q_j=q_{2j-1}+iq_{2j}$ and $k_j=k_{2j-1}+ik_{2j}$. Then
	\begin{equation}
		q_j\mapsto e^{i\omega_jp}q_j,
		\qquad
		k_j\mapsto e^{i\omega_jp}k_j,
		\label{eq:rope-complex-action}
	\end{equation}
	and the real attention logit is
	\begin{equation}
		\ell_{t,s}
		=\sum_{j=1}^{F}
		\operatorname{Re}\!\left[
		\overline{q_{t,j}}\,
		e^{i\omega_j(s-t)}
		k_{s,j}
		\right].
		\label{eq:rope-complex-logit}
	\end{equation}
	The complex notation is only a compact representation of ordinary real
	$2\times2$ RoPE pairs; deployment remains real-valued. Each RoPE frequency
	induces a distinct positional rotation. A static latent coordinate that mixes
	different frequencies therefore cannot itself follow one original RoPE
	rotation.
	\rev{In multi-head and grouped-query attention, copies of the same frequency across heads rotate identically, providing multiple coordinates that can be compressed while preserving their shared positional action.}
	
	\subsection{Why Generic Latent Keys Remain Reconstructive}
	\label{sec:bg-lowrank}
	
	Consider a generic low-rank key representation with column-vector hidden state
	$\mathbf{x}_s\in\mathbb{R}^{d}$. Using the conventional projection shape
	$\mathbf{W}_K\in\mathbb{R}^{d\times d_K}$, so that $\mathbf{k}_s=\mathbf{W}_K^\top \mathbf{x}_s$, write
	\begin{equation}
		\mathbf{W}_K\approx \mathbf{A}_K\mathbf{B}_K,
		\qquad
		\mathbf{A}_K\in\mathbb{R}^{d\times r_K},
		\qquad
		\mathbf{B}_K\in\mathbb{R}^{r_K\times d_K},
		\label{eq:palu-factorization}
	\end{equation}
	and cache the column-vector latent
	\begin{equation}
		\mathbf{h}_s^K=\mathbf{A}_K^\top \mathbf{x}_s,
		\qquad
		\hat{\mathbf{k}}_s=\mathbf{B}_K^\top \mathbf{h}_s^K.
		\label{eq:palu-key-cache}
	\end{equation}
	Palu~\citep{palu} is a representative post-training construction of this form.
	
	Without a position-dependent transformation, the decoder can be absorbed into
	the query side:
	\begin{equation}
		\mathbf{q}_t^\top\hat{\mathbf{k}}_s
		=\mathbf{q}_t^\top \mathbf{B}_K^\top \mathbf{h}_s^K
		=\bigl(\mathbf{B}_K\mathbf{q}_t\bigr)^\top \mathbf{h}_s^K.
		\label{eq:nope-absorb}
	\end{equation}
	The value path has the same useful linearity. If
	$\mathbf{h}_s^V=\mathbf{A}_V^\top \mathbf{x}_s$ and $\hat{\mathbf{v}}_s=\mathbf{B}_V^\top \mathbf{h}_s^V$, then
	\begin{equation}
		\sum_s\alpha_{t,s}\hat{\mathbf{v}}_s
		=\mathbf{B}_V^\top\left(\sum_s\alpha_{t,s}\mathbf{h}_s^V\right),
		\label{eq:value-absorb}
	\end{equation}
	so value decoding can occur after aggregation.
	
	RoPE breaks the analogous absorption on the key path. The score becomes
	\begin{equation}
		(\mathbf{R}_t\mathbf{q}_t)^\top \mathbf{R}_s\mathbf{B}_K^\top \mathbf{h}_s^K,
		\label{eq:palu-rope-score}
	\end{equation}
	\rev{Moving the decoder to the query side yields $\mathbf{B}_K\mathbf{R}_s^\top\mathbf{R}_t\mathbf{q}_t$, which depends on the historical position $s$. For a generic decoder, a single query projection therefore cannot serve the entire history.}
	Implementations may fuse reconstruction, rotation, and scoring without
	writing the full key to off-chip memory, yet reconstruction remains part of the
	rotary key path.

	\begin{revision}
	Direct execution requires a compatible positional action in latent space.
	Writing $\mathbf{U}=\mathbf{B}_K^\top$, we seek original-frequency latent rotations
	$\mathbf{R}'_p$ satisfying Equation~(\ref{eq:intro-intertwine}).
	This relation makes rotating a latent key before decoding equivalent
	to rotating its decoded key in native space. Section~\ref{sec:theorysec}
	characterizes the static decoders satisfying this condition and connects it to
	direct latent scoring.
	\end{revision}
	
	\section{RoPE-Compatible Latent Spaces}
	\label{sec:theorysec}
	
	\paragraph{Same-frequency blocks.}
	We characterize static latent spaces on which pretrained RoPE remains closed.
	For frequency $j$, stack the pre-RoPE key coordinates from all heads that share
	that frequency as $\boldsymbol{\kappa}_j\in\mathbb{C}^D$; thus $D$ simply counts the
	same-frequency copies. In GQA~\citep{gqa_ainslie}, $D=H_{kv}$; in MHA,
	$D=H_q=H_{kv}$. All $D$ coordinates receive the same RoPE
	phase $e^{i\omega_jp}$, and a rank-$r_j$ latent block uses decoder
	$\mathbf{U}_j\in\mathbb{C}^{D\times r_j}$.
	
	\begin{theorem}[Exact RoPE-compatible maps]
		\label{thm:intertwiner}
		Assume the retained RoPE frequencies are pairwise distinct and no retained
		frequency equals the negative of another modulo $2\pi$:
		$\omega_j\not\equiv\omega_k\pmod{2\pi}$ for $j\neq k$ and
		$\omega_j\not\equiv-\omega_k\pmod{2\pi}$ for all $j,k$. Let each latent
		block use its corresponding original RoPE frequency. \rev{A static real-linear
		decoder is exactly RoPE-compatible if and only if it is block diagonal in
		frequency and each same-frequency block is the realification of a
		complex-linear map.}
	\end{theorem}
	
	\begin{proof}[Sketch]
		\rev{Decompose each real-linear block as $Tz=Az+B\bar z$.
		Intertwining forces $A$ to vanish between distinct frequencies and $B$ to
		vanish unless the frequencies are negatives. The assumptions exclude
		all such antilinear terms; same-frequency complex-linear blocks commute
		with the shared rotary phase.}
		Appendix~\ref{app:proofs} gives the full proof.
	\end{proof}
	
	\textit{Operationally, exact RoPE-compatible compression may mix across heads sharing a frequency, but cannot mix coordinates with different RoPE frequencies.}

	The theorem constrains executability rather than reconstruction quality.
	Within this compatible family, fidelity depends on retained rank and residual
	correction; increasing $r_j$ adds capacity without changing the positional
	rule. Appendix~\ref{app:finite-window} discusses finite-window near-degeneracy.
	
	\textbf{From compatibility to direct execution.}
	To connect the closure condition to attention, consider a semi-unitary
	compatible decoder $\mathbf{U}$ with $\mathbf{U}^\dagger \mathbf{U}=\mathbf{I}$. Its adjoint provides the
	corresponding projection into latent coordinates. A historical token at
	position $s$ has latent coordinates $\mathbf{z}_s=\mathbf{U}^\dagger \mathbf{k}_s$. The score
	against the compressed reconstruction $\mathbf{U}\mathbf{U}^\dagger \mathbf{k}_s$ is
	\begin{equation}
		\langle \mathbf{R}_t\mathbf{q}_t,\mathbf{R}_s\mathbf{U}\mathbf{U}^\dagger \mathbf{k}_s\rangle_{\mathbb{R}}
		=
		\langle \mathbf{R}'_t\mathbf{U}^\dagger \mathbf{q}_t,\mathbf{R}'_s\mathbf{z}_s\rangle_{\mathbb{R}}
		=
		\langle \mathbf{U}^\dagger \mathbf{q}_t,\mathbf{R}'_{s-t}\mathbf{z}_s\rangle_{\mathbb{R}}.
		\label{eq:direct-latent-score}
	\end{equation}
	\rev{The implementation caches $\mathbf c_s=\mathbf R'_s\mathbf z_s$ and scores it against $\mathbf R'_t\mathbf U^\dagger\mathbf q_t$, without native-key reconstruction.}
	
	\begin{corollary}[Direct latent scoring]
		\label{cor:absorb}
		Under static linear encoder, decoder, and query-side maps, with the latent
		space taken to be the full image of the encoder, original-frequency latent
		rotations, and exact logit equality relative to the compressed reconstruction,
		exact direct latent scoring is possible if
		and only if the key decoder intertwines RoPE:
		\begin{equation}
			\mathbf{R}_p\mathbf{U}=\mathbf{U}\mathbf{R}'_p
			\qquad\text{for all }p.
		\end{equation}
	\end{corollary}
	Equation~(\ref{eq:direct-latent-score}) gives sufficiency and
	Appendix~\ref{app:proofs} proves the converse. Thus, under this execution model,
	intertwining is the structural condition for a reconstruction-free fixed
	latent-RoPE execution path.
	
	\begin{proposition}[Rank-one phase-gauge equivalence]
		\label{prop:gauge}
		Every complex rank-one RoPE-compatible carrier admits an exactly equivalent
		phase-aligned real realization by folding per-head phases into the
		pretrained query and key projections. This reparameterization preserves the pretrained attention function, and the ungauged real rank-one family is a
		strict subset of the phase-aligned family.
	\end{proposition}
	Operationally, phase alignment bridges the complex-linear characterization
	and real-valued inference: it implements the full complex rank-one compatible
	family with real operators without increasing compatible rank. Appendix~\ref{app:proofs} gives the construction.

	\section{GaugeRoPE}
	\label{sec:method}
	
	\begin{revision}
	The compatibility characterization determines which latent maps preserve pretrained RoPE; constructing a compact cache additionally requires choosing what information to retain. GaugeRoPE selects a low-rank carrier within each compatible frequency block to capture rotary key information. Since a finite-rank carrier leaves part of the key unrepresented, a joint residual key--value (RKV) code stores complementary key content together with values. The carrier retains its original rotary action, while the residual correction uses position-independent absorbed maps so that both branches remain directly executable.

	\paragraph{Construction and execution.}
	Carrier bases and RKV maps are fitted offline from calibration activations, with the pretrained model weights frozen. Each incoming token is encoded into carrier coordinates and an RKV code. During decoding, the query is mapped into these latent spaces to score the cached history, and values are aggregated before decoding the output. This construction moves the decoder maps outside the historical token traversal, avoiding per-token native-dimensional K/V reconstruction.
	\end{revision}

	\subsection{RoPE-Compatible Carrier Construction}
	\label{sec:carrier}
	\label{sec:compatible-rank}
	
	For one layer and frequency $j$, let
	$\boldsymbol{\kappa}_j(t)\in\mathbb{C}^{D}$ stack the $D$ pre-RoPE key coordinates
	sharing that RoPE frequency at calibration token $t$ ($D=H_{kv}$ for GQA
	and $D=H_q=H_{kv}$ for MHA). We form the same-frequency Gram matrix
	\begin{equation}
		\mathbf{G}_j=\sum_t\boldsymbol{\kappa}_j(t)\boldsymbol{\kappa}_j(t)^\dagger
		\in\mathbb{C}^{D\times D},
		\label{eq:gram}
	\end{equation}
	and use its leading eigenvectors as the carrier basis $\mathbf{U}_j$. The
	eigendecomposition is performed independently within the same-frequency blocks
	allowed by Theorem~\ref{thm:intertwiner}. Consequently, changing the retained
	rank changes approximation capacity but not the RoPE action carried by each
	latent coordinate.
	
	For a rank-one basis, write $\mathbf{u}_j=\mathbf{D}_j\mathbf{a}_j$, where $\mathbf{D}_j$ is diagonal with the
	unit phases of $\mathbf{u}_j$ (assigning an arbitrary unit phase to zero entries) and
	$\mathbf{a}_j=|\mathbf{u}_j|$ is real. Then
	\begin{equation}
		\mathbf{u}_j^\dagger\boldsymbol{\kappa}_j=\mathbf{a}_j^\top \mathbf{D}_j^\dagger\boldsymbol{\kappa}_j.
		\label{eq:phase-realization}
	\end{equation}
	We fold $\mathbf{D}_j^\dagger$ into the Q/K projections offline using
	Proposition~\ref{prop:gauge}, leaving a real projection at inference. This
	reparameterization preserves the pretrained attention function. \rev{Higher ranks, including R2, implement complex coefficients through real $2\times2$ blocks (Appendix~\ref{app:gqa-realization}).}
	
	Here rank means the number of retained directions within a same-frequency
	block, not attention-operator rank. For $D=1$, rank one already spans the full
	frequency block; for $D>1$, Theorem~\ref{thm:intertwiner} allows additional
	modes:
	\begin{equation}
		\mathbf{U}_j=[\mathbf{u}_{j,1},\mathbf{u}_{j,2},\ldots,\mathbf{u}_{j,r_j}],
		\qquad
		\mathbf{z}_{s,j}=\mathbf{U}_j^\dagger\boldsymbol{\kappa}_{s,j}.
		\label{eq:multirank}
	\end{equation}
	Every added column remains in the same-frequency block and therefore keeps
	the same RoPE frequency. In the main GQA configuration, we use a
	uniform rank across frequencies: $r_j=1$ for R1 and $r_j=2$ for R2. For
	Mistral, $D=8$, so these settings retain one or two of eight available
	same-frequency modes; the MHA experiments instead compare matched normalized
	ranks \rev{$\alpha=r/D$}. The theorem also permits frequency-dependent ranks $r_j$, but we
	use the uniform construction throughout the main experiments. For head
	dimension $d_h$, each compatible mode contributes exactly $d_h$ real
	coordinates per layer and token.
	
	\begin{figure*}[t]
		\centering
		\includegraphics[width=0.90\textwidth]{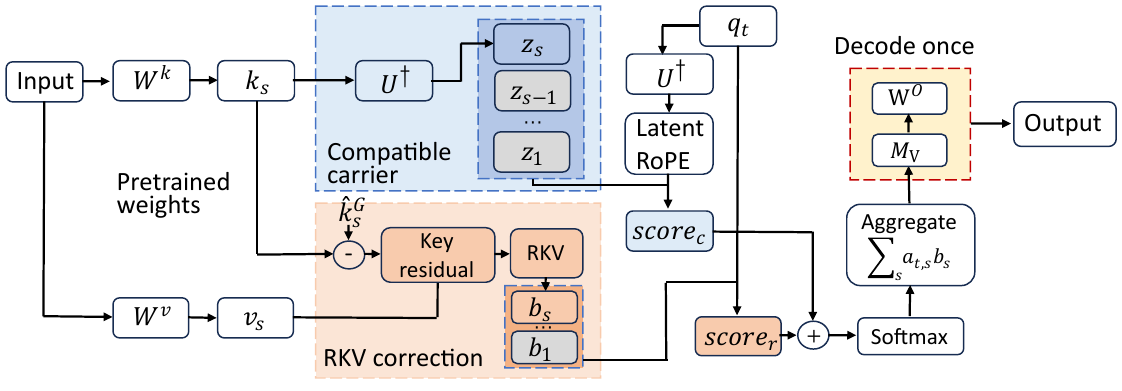}
		\caption{
			\textbf{GaugeRoPE overview.}
			GaugeRoPE represents each historical token with two latent
			components: a RoPE-compatible carrier $\mathbf{z}_s$ for the rotary key path
			and a position-independent RKV code $\mathbf{b}_s$ for complementary residual-key
			and value information. The two paths produce complementary
			attention scores that are combined before softmax, while values are
			aggregated in latent space and decoded once, avoiding per-token historical
			native-K/V reconstruction.
		}
		\label{fig:overview}
	\end{figure*}
	
	\subsection{Residual Key-Value Correction and Direct Attention}
	\label{sec:content-latent}
	\label{sec:direct-execution}
	
	The compatible carrier need not span the entire key. As in decoupled-RoPE latent attention~\citep{deepseekv2}, explicit positional interaction can be concentrated in a dedicated subspace, leaving the remaining capacity for residual key and value content. Let $\hat{\mathbf{k}}^{\mathrm{G}}$ denote the key component reconstructed from the compatible carrier and define
	\begin{equation}
		\mathbf{r}^K=\mathbf{k}-\hat{\mathbf{k}}^{\mathrm{G}}.
		\label{eq:key-residual}
	\end{equation}
	\rev{This residual is used to fit the RKV encoder offline and to encode each token when it enters the cache.}
	For the RKV latent, $\mathbf{r}^K$ and $\mathbf{v}$ are flattened over the $H_{kv}$ heads, each
	in $\mathbb{R}^{H_{kv}d_h}$. After energy balancing, a joint PCA encoder
	$\mathbf{A}_R$ produces
	\begin{equation}
		\mathbf{b}_s=\mathbf{A}_R^\top[\mathbf{r}_s^K;\mathbf{v}_s],
		\qquad
		\hat{\mathbf{r}}_s^K=\mathbf{M}_K\mathbf{b}_s,
		\qquad
		\hat{\mathbf{v}}_s=\mathbf{M}_V\mathbf{b}_s,
		\label{eq:rkv-codec}
	\end{equation}
	where $\mathbf{M}_K$ and $\mathbf{M}_V$ are the residual-key and value blocks of the decoder.
	For query head $g$, let $h=\pi(g)$ be its shared KV head. Writing
	$\mathbf{e}_h\in\mathbb{C}^D$ for the corresponding coordinate in each same-frequency
	block, define the absorbed carrier query
	$\widetilde{\mathbf{q}}_{t,g,j}=\mathbf{U}_j^\dagger \mathbf{e}_h q_{t,g,j}$. The carrier score is
	\begin{equation}
		\ell^{\mathrm G}_{t,s,g}
		=\sum_j \operatorname{Re}\!\left[
		\widetilde{\mathbf{q}}_{t,g,j}^\dagger
		e^{i\omega_j(s-t)} \mathbf{z}_{s,j}
		\right],
		\label{eq:carrier-score-head}
	\end{equation}
	which is the per-head GQA realization of Equation~(\ref{eq:direct-latent-score}).
	For the RKV correction, let
	$\mathbf{S}_h\in\mathbb{R}^{d_h\times H_{kv}d_h}$ select the same KV-head block. The
	residual-key decoder is absorbed into the query side,
	\begin{equation}
		\ell^{\mathrm{res}}_{t,s,g}
		=\mathbf{q}_{t,g}^\top \mathbf{S}_h\mathbf{M}_K\mathbf{b}_s
		=\bigl(\mathbf{M}_K^\top \mathbf{S}_h^\top \mathbf{q}_{t,g}\bigr)^\top \mathbf{b}_s.
		\label{eq:rkv-residual-score}
	\end{equation}
	Historical scoring for query head $g$ is therefore
	\begin{equation}
		\hat\ell_{t,s,g}
		=
		\underbrace{\ell^{\mathrm G}_{t,s,g}}_{\text{exact RoPE-compatible carrier}}
		+
		\underbrace{\ell^{\mathrm{res}}_{t,s,g}}_{\text{position-independent correction}}.
		\label{eq:score-decomposition}
	\end{equation}
	\rev{Attention weights use $\operatorname{softmax}(\hat\ell/\sqrt{d_h})$ over causally available keys.} The carrier score is exact within the retained subspace. The RKV branch
	approximates the remaining position-dependent score
	$\mathbf{q}_{t,g}^\top \mathbf{R}_{s-t}\mathbf{S}_h\mathbf{r}_s^K$ by the position-independent
	$\mathbf{q}_{t,g}^\top \mathbf{S}_h\hat{\mathbf{r}}_s^K$, enabling query-side decoder absorption.
	Thus the overall attention is approximate even though carrier execution is exact.
	
	Figure~\ref{fig:overview} shows the hybrid path: the carrier is scored in
	RoPE-compatible latent coordinates and the RKV correction is scored directly
	from its position-independent code before softmax. Values admit the same
	head-selective linear absorption. For query head $g$ with $h=\pi(g)$,
	\begin{equation}
		\sum_s\alpha_{t,s,g}\mathbf{S}_h\hat{\mathbf{v}}_s
		=\mathbf{S}_h\mathbf{M}_V\left(\sum_s\alpha_{t,s,g}\mathbf{b}_s\right).
		\label{eq:latent-value-aggregation}
	\end{equation}
	Both paths therefore avoid per-token historical native K/V reconstruction.

	\subsection{State Size}
	\label{sec:cache-layout}
	
	The theory fixes representation structure but leaves the retained capacity as
	a design choice, so experiments compare fixed state budgets. With
	per-frequency compatible rank $r$ and a $d_c$-dimensional RKV code, the
	\rev{KV cache contains}
	\begin{equation}
		d_hr+d_c
		\label{eq:gauge-real-dims}
	\end{equation}
	real scalars per layer and token. A full GQA KV state contains
	$2H_{kv}d_h$ scalars, so the nominal dimensional ratio is
	\begin{equation}
		\rho
		=
		\frac{2H_{kv}d_h}{d_hr+d_c}.
		\label{eq:gauge-compression}
	\end{equation}
	\begin{revision}
	With carrier and RKV bit widths $b_{\mathrm G}$ and $b_{\mathrm R}$, the cache uses
	\begin{equation}
		B_{\mathrm{cache}}=\frac{b_{\mathrm G}d_hr+b_{\mathrm R}d_c}{8}+B_{\mathrm{meta}}
		\label{eq:gauge-cache-bytes}
	\end{equation}
	bytes per layer and token, where $B_{\mathrm{meta}}$ accounts for token-dependent
	quantization metadata. Shared static scales are counted separately.
	\end{revision}

\section{Experiments}
\label{sec:experiments}

\subsection{Experimental Setup}
	\label{sec:exp-setup}
	
	\textbf{Models and budgets.}
	We use Mistral-7B-Instruct-v0.3 as the primary GQA backbone for \rev{quality comparisons,
	mechanistic analysis, and system evaluation}. \rev{Main Mistral comparisons use $2\times$ low-rank compression and quantized configurations: $5.20\times$/$5.24\times$ for Palu/ReCalKV with 6-bit latents, and $5.85\times$ for GaugeRoPE with 8-bit carrier and 4-bit RKV storage. Ratios are relative to Full's BF16 KV cache. We abbreviate GaugeRoPE as Gauge in tables and figures. The suffix -50\% denotes a 50\% reduction in cached dimensions; bit widths indicate quantization, with 16-bit storage used otherwise. GaugeRoPE uses R2 with $d_c=768$, totaling 1024 real scalars per token/layer (Section~\ref{sec:cache-layout}); Appendix~\ref{app:quantization} gives calibration and storage details. Qwen2.5-7B provides an additional GQA backbone.} \rev{CodeLlama-7B-Instruct-hf tests the same construction on standard MHA.}
	
	\textbf{Baselines.}
	\textbf{Full} denotes the unmodified pretrained model.
	\textbf{Palu}~\citep{palu} and \textbf{ReCalKV}~\citep{recalkv} are both
	post-training low-rank methods. \rev{Their 50\% configurations match GaugeRoPE-50\% in per-token state budget. \textbf{KIVI}~\citep{kivi} provides a direct KV-quantization baseline using its official 2-bit implementation in FP16. Its compression ratios average measured cache reduction across examples, including metadata and residual storage.}
	
	\textbf{Benchmarks.}
	RULER~\citep{ruler} is evaluated on all 13 tasks with 50 examples
	per task and the official metric. Mistral is additionally evaluated on all 16
	English LongBench tasks~\citep{longbench}, with native lengths capped at 32K
	and official task-specific metrics. \rev{All methods use the same 50 LongBench examples per task, scored in the same environment.}

\subsection{Long-Context Quality and Generalization}
	\label{sec:mistral-main}
	
	\rev{Tables~\ref{tab:ruler-category-main} and~\ref{tab:longbench-category-main} compare Mistral quality under low-rank compression and quantization. GaugeRoPE-50\% (8/4 bits) achieves $5.85\times$ compression with 85.56 RULER and 35.73 LongBench, closely matching the 85.92/35.93 scores of its unquantized counterpart. Quantization reduces per-token KV storage by 65.8\% with only a 0.20-point decrease on LongBench. Compared with the 6-bit Palu and ReCalKV configurations, GaugeRoPE uses 11.2\% and 10.4\% less per-token KV storage while improving RULER by 2.64 and 4.64 points. Palu and ReCalKV score 37.88 and 37.61 on LongBench, with different strengths across task categories.}
	
	\begin{table}[htbp]
		\centering
		\footnotesize
		\renewcommand{\arraystretch}{1.08}
		\setlength{\tabcolsep}{11.7pt}
		\caption{Mistral-7B-Instruct-v0.3 RULER category macros. For GaugeRoPE, 8/4 bits denotes 8-bit carrier and 4-bit RKV storage. NIAH averages the eight needle-in-a-haystack tasks;
			CWE--FWE and QA average two tasks each; Avg. averages all 13 tasks.}
		\label{tab:ruler-category-main}
		\begin{tabular}{@{}>{\columncolor{white}[0pt][\tabcolsep]}lrrrrr>{\columncolor{white}[\tabcolsep][0pt]}r@{}}
			\toprule
			Method & Comp. Ratio & NIAH & VT & CWE--FWE & QA & Avg. \\
			\midrule
			Full        & $1.00\times$ & 95.00 & 98.80 & 93.07 & 68.00 & 90.84 \\
			\midrule
			Palu-50\%   & $2.00\times$ & 91.19 & 92.80 & 75.50 & 55.00 & 83.33 \\
			ReCalKV-50\% & $2.00\times$ & 88.44 & 92.80 & 74.83 & 55.00 & 81.54 \\
			\rowcolor{gaugeRow}
			Gauge-50\%  & $2.00\times$ & 91.13 & 90.80 & 82.60 & 66.00 & 85.92 \\
			\midrule
			KIVI (2 bits) & $5.06\times$ & 85.63 & 97.60 & 91.83 & 67.00 & 84.64 \\
			Palu-50\% (6 bits) & $5.20\times$ & 90.13 & 93.20 & 76.87 & 55.00 & 82.92 \\
			ReCalKV-50\% (6 bits) & $5.24\times$ & 88.38 & 92.00 & 73.47 & 53.00 & 80.92 \\
			\rowcolor{gaugeRow}
			Gauge-50\% (8/4 bits) & $5.85\times$ & 91.31 & 92.00 & 81.87 & 63.00 & 85.56 \\
			\rowcolor{gaugeRow}
			\quad + 128 recent tokens & $5.42\times$ & 91.88 & 92.80 & 86.13 & 61.00 & 86.31 \\
			\bottomrule
		\end{tabular}
	\end{table}
	
	\begin{table}[htbp]
		\centering
		\footnotesize
		\renewcommand{\arraystretch}{1.08}
		\setlength{\tabcolsep}{3.6pt}
		\caption{Mistral-7B-Instruct-v0.3 LongBench results on the same 800 examples: 16 English tasks with 50 examples each and a 32K input cap. Configurations match Table~\ref{tab:ruler-category-main}; Avg. is the task-equal macro.}
		\label{tab:longbench-category-main}
		\begin{tabular}{@{}>{\columncolor{white}[0pt][\tabcolsep]}lrrrrrrr>{\columncolor{white}[\tabcolsep][0pt]}r@{}}
			\toprule
			Method & Comp. Ratio & \shortstack{Single-doc\\QA} & \shortstack{Multi-doc\\QA} & Summ. & Few-shot & Synthetic & Code & Avg. \\
			\midrule
			Full        & $1.00\times$ & 40.28 & 36.91 & 29.57 & 47.29 & 51.00 & 58.68 & 42.59 \\
			\midrule
			Palu-50\%   & $2.00\times$ & 39.00 & 36.50 & 26.39 & 49.37 & 28.00 & 49.13 & 38.00 \\
			ReCalKV-50\% & $2.00\times$ & 37.55 & 30.66 & 26.50 & 44.28 & 36.00 & 49.48 & 36.74 \\
			\rowcolor{gaugeRow}
			Gauge-50\%  & $2.00\times$ & 32.23 & 32.07 & 28.35 & 35.11 & 41.00 & 54.83 & 35.93 \\
			\midrule
			KIVI (2 bits) & $5.05\times$ & 39.86 & 37.90 & 28.78 & 44.04 & 51.00 & 57.31 & 41.77 \\
			Palu-50\% (6 bits) & $5.20\times$ & 39.64 & 36.83 & 26.80 & 49.82 & 25.00 & 48.42 & 37.88 \\
			ReCalKV-50\% (6 bits) & $5.24\times$ & 37.46 & 32.13 & 26.57 & 45.65 & 38.00 & 50.19 & 37.61 \\
			\rowcolor{gaugeRow}
			Gauge-50\% (8/4 bits)  & $5.85\times$ & 32.39 & 30.36 & 27.76 & 34.49 & 45.00 & 53.33 & 35.73 \\
			\rowcolor{gaugeRow}
			\quad + 128 recent tokens & $5.41\times$ & 34.79 & 38.68 & 28.59 & 37.81 & 44.00 & 55.26 & 38.63 \\
			\bottomrule
		\end{tabular}
	\end{table}

\begin{revision}
The advantage extends beyond needle retrieval: the mixed-precision configuration scores 63.00 on RULER QA versus 55.00 and 53.00 for 6-bit Palu and ReCalKV. On LongBench, GaugeRoPE leads both low-rank baselines in summarization, synthetic tasks, and code; its passage-retrieval score is 88.00, compared with 48.00 and 74.00. Palu and ReCalKV perform better on document QA and few-shot tasks. Compared with KIVI, GaugeRoPE offers higher compression and stronger NIAH retrieval (91.31 vs.\ 85.63), while KIVI achieves higher LongBench quality (41.77 vs.\ 35.73).

Retaining 128 recent tokens in native K/V further raises GaugeRoPE's scores to 86.31 on RULER and 38.63 on LongBench, at mean measured compression ratios of $5.42\times$ and $5.41\times$ including the native cache. The LongBench gain is 2.90 points (paired 95\% CI $[1.50,4.34]$).

\end{revision}

\begin{revision}
\noindent
\begin{minipage}[t]{.49\linewidth}
\vspace{0pt}
\textbf{Backbone generalization.}\label{sec:mha-generality}
Table~\ref{tab:backbone-main} tests the same construction across GQA and standard MHA. GaugeRoPE retains 88.1\% of Full's RULER score on Qwen2.5-7B. On CodeLlama, the larger head multiplicity supports a wider compatible carrier under the same construction.
\end{minipage}\hfill
\begin{minipage}[t]{.48\linewidth}
\vspace{0pt}\color{black}\centering
\captionof{table}{\textbf{Cross-backbone quality.} RULER macros with BF16 caches at nominal $2\times$ compression.}
\footnotesize
\setlength{\tabcolsep}{3pt}
\label{tab:backbone-main}
\begin{tabular*}{\linewidth}{@{\extracolsep{\fill}}llrr@{}}
\toprule
Backbone & Attention & Full & Gauge \\
\midrule
Qwen2.5-7B & GQA & 89.63 & 78.96 \\
CodeLlama-7B & MHA & 79.57 & 69.50 \\
\bottomrule
\end{tabular*}
\end{minipage}
\par\medskip
On CodeLlama, increasing the carrier fraction $\alpha$ from $1/8$ to $1/4$ improves RULER by 1.72 points at the same cache budget. Length scaling further demonstrates GaugeRoPE's ability to preserve long-range retrieval. On the four-task RULER NIAH subset, it retains 88.6\% and 84.6\% of Full's score at 32K on Mistral and Qwen2.5-7B, respectively. On Qwen2.5-7B, its advantage over TransMLA at the same cache budget grows with context length, with neither method using uptraining. Appendices~\ref{app:mha-generality} and~\ref{app:additional-quality} give the rank sweep, length scaling, and evaluation configurations.
\end{revision}

\begin{revision}
\subsection{Compatible Capacity and Attention Fidelity}
\label{sec:ablations}
\begin{figure}[!htbp]
\centering
\includegraphics[width=\linewidth]{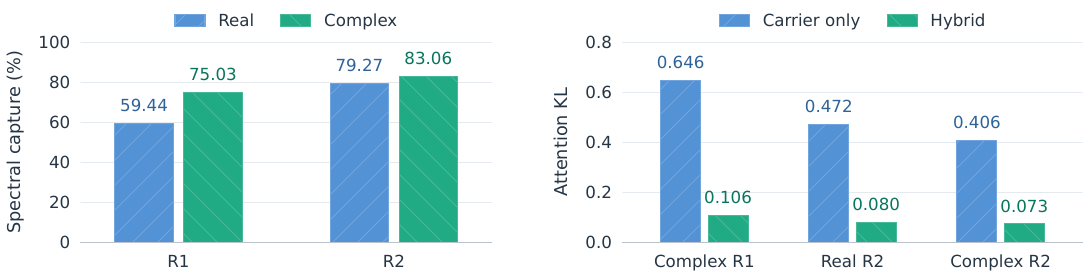}
\caption{\textbf{Compatible capacity and coefficient freedom.} Left: same-frequency Gram energy captured by real and complex carriers. Right: attention KL on identical normalized Full states at a fixed budget of 1024 real scalars per token/layer.}
\label{fig:mechanisms}
\end{figure}
The first design choice is how much of a fixed cache budget to assign to the compatible carrier. We compare one and two retained modes per frequency (R1 and R2), which preserve the same positional action but differ in capacity. R2 assigns more dimensions to the carrier and fewer to RKV, yet produces attention distributions closer to Full (Figure~\ref{fig:mechanisms}, right). With RKV refitted for each carrier, RULER rises from 6.36 to 85.92. The improvement despite a smaller residual code shows the value of retaining more key information in the rotary branch.

At a given rank, coefficient freedom determines how effectively the carrier uses its capacity. Real coefficients mix heads with the same weights for both coordinates of each RoPE pair; complex coefficients additionally allow relative phase shifts. Compared with a RoRoPE-style real carrier at the same rank and native frequencies, the complex carrier captures more key energy (Figure~\ref{fig:mechanisms}, left). At R2, this improves attention fidelity without increasing cache size.

The improved fidelity is accompanied by more stable generation across three calibration samples: Complex R2 scores 78.00--84.75 on the four-task RULER subset, while Real R2 produces repetitive continuations and scores zero in every run. Layerwise diagnostics help explain this gap: complex coefficients reduce first-layer KL by 56.5\%, compared with approximately 10\% averaged across layers. On two traced prompts, the hidden-state trajectories already diverge during prefill, while direct execution agrees numerically with dense reconstruction. Together, the larger early-layer difference and divergence before generation suggest that errors accumulate during prefill, helping explain the large generation gap despite a modest change in average KL (Appendices~\ref{app:complex-control} and~\ref{app:sample-sensitivity}).

The carrier's finite capacity also explains the importance of the residual branch. Adding the RKV correction to R2 lowers attention KL from 0.406 to 0.073 by recovering key information omitted by the carrier. GaugeRoPE evaluates this correction directly from the cached code without RoPE. Restoring RoPE to the same compressed residual lowers KL further to 0.025, but requires residual-key reconstruction. This controlled comparison isolates the positional approximation that enables direct execution of the residual branch (Appendix~\ref{app:residual-attribution}).
\end{revision}

\begin{revision}
\subsection{Efficient Long-Context Decoding}
\label{sec:system-exp}
\label{sec:direct-materialized}
GaugeRoPE removes historical KV reconstruction from decoding. Our dedicated Triton split-PV kernel fuses attention-weight computation with latent-value aggregation and uses incremental cache updates. Decoder maps act on the current query and aggregated output, keeping historical attention in compressed space. We compare against Full with the PyTorch FlashAttention backend and Palu/ReCalKV implementations based on Palu's official fused kernels. All four methods use BF16 caches, with the low-rank methods retaining 50\% of Full's historical KV dimensions. Appendix~\ref{app:system-protocol} gives implementation and measurement details.

\begin{figure}[!htbp]
\centering
\includegraphics[width=\linewidth]{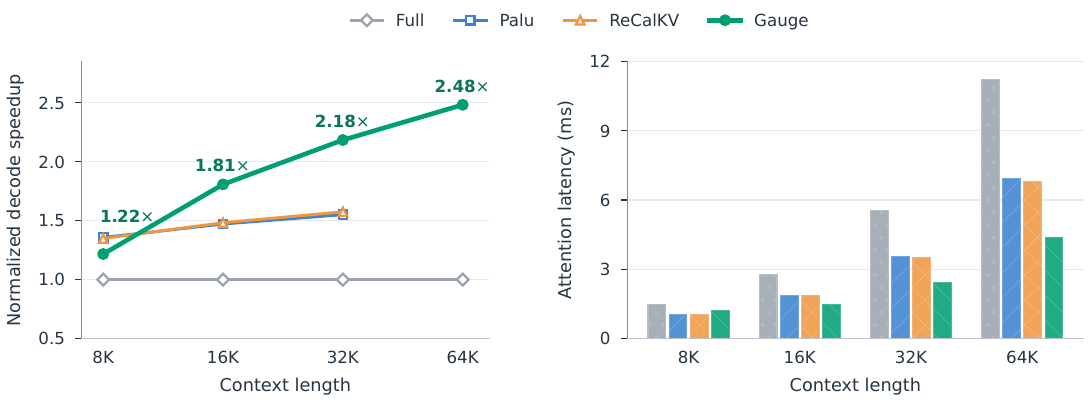}
\caption{\textbf{Decoding at batch size 4 with BF16 caches.} Left: full-model throughput normalized to Full; Palu/ReCalKV run out of memory during initialization at 64K. Right: attention latency averaged over 32 layers, measured one layer at a time.}
\label{fig:e2e-throughput}
\end{figure}

In full-model decoding, GaugeRoPE overtakes both compressed baselines from 16K onward (Figure~\ref{fig:e2e-throughput}, left). At 32K, it delivers $2.18\times$ Full's throughput and exceeds Palu and ReCalKV by 40.8\% and 38.7\%, respectively; at 64K, its advantage over Full reaches $2.48\times$. Longer histories increase the reconstruction work avoided at each step, making direct execution increasingly valuable.

Layerwise measurements include projections, attention, output mapping, and cache updates (Figure~\ref{fig:e2e-throughput}, right). At 32K, GaugeRoPE reduces latency by 32.0\% and 31.3\% relative to Palu and ReCalKV; at 64K, the reductions reach 36.8\% and 35.7\%. Palu's fused implementation avoids writing reconstructed keys to GPU global memory by combining reconstruction, RoPE, and scoring. GaugeRoPE additionally removes the reconstruction arithmetic, translating the compatibility property in Section~\ref{sec:theorysec} into faster attention.

For unquantized caches, GaugeRoPE reduces persistent KV storage from Full's 32.06 GiB to 16.34 GiB at 64K. Its additional codec storage is 1.127 GiB, compared with 3.039 GiB for Palu and 6.055 GiB for ReCalKV, contributing to lower full-model peak memory (Appendix~\ref{app:system-protocol}). This unquantized implementation also supports larger batches and Qwen2.5-14B-Instruct (Table~\ref{tab:scalability}). At 16K, Mistral's speedup over Full grows from $2.31\times$ at batch 8 to $2.51\times$ at batch 16; Qwen2.5-14B-Instruct achieves $2.26\times$ at batch 4 and 32K.
\end{revision}

\FloatBarrier

	\section{Related Work}
	\label{sec:related}
	
	\textbf{Latent attention and architecture conversion.}
	MLA~\citep{deepseekv2,deepseekv3} learns a compressed content latent together
	with a separate RoPE branch during pretraining. TransMLA~\citep{transmla}
	and MHA2MLA~\citep{mha2mla} adapt pretrained attention toward MLA-style
	representations. \rev{TransMLA's RoRoPE applies the same real orthogonal mixing to real and imaginary coordinates within each frequency, retains principal rotary modes, and removes RoPE from the remaining components before balanced joint KV PCA. GaugeRoPE extends same-frequency real mixing to the full complex-linear family and establishes the conditions for direct latent execution under pretrained RoPE. RoRoPE's real carrier is the special case $\mathbf B_j=0$ of $\mathbf U_j=\mathbf A_j+i\mathbf B_j$.}
	
	\textbf{Low-rank and structured KV compression.}
	Palu~\citep{palu} is our primary training-free low-rank reference: its
	RoPE-aware key path reconstructs native-dimensional historical keys before
	positional scoring. Eigen Attention~\citep{eigenattention} obtains per-head
	low-rank Q/K subspaces by SVD without RoPE-frequency constraints and likewise
	returns compressed keys to native dimension before rotation. ReCalKV~\citep{recalkv},
	StiefAttention~\citep{stiefattention}, STAR-KV~\citep{starkv},
	ShadowKV~\citep{shadowkv}, and xKV~\citep{xkv} improve low-rank fidelity or
	systems efficiency through calibration, learned projectors, reconstruction,
	sparse selection, or cross-layer redundancy. GaugeRoPE targets direct executability under the pretrained RoPE operator, allowing historical attention to be evaluated from cached latent states without native-key reconstruction.
	
	\textbf{RoPE-aware and complementary cache compression.}
	RAP~\citep{rap} preserves RoPE coordinate pairs through structured pruning.
	From our perspective, this corresponds to coordinate selection within the
	compatible family, whereas our characterization additionally permits arbitrary
	complex-linear mixing across equal-frequency head copies.
	CommVQ~\citep{commvq} instead enforces a commuting codebook interface,
	addressing a discrete quantization setting outside our static-linear
	characterization. SALS~\citep{sals} targets the same RoPE-induced reconstruction
	bottleneck, but uses RoPE-free query--key interactions in latent space to select
	a sparse subset of tokens and reconstructs only the selected states, rather than
	maintaining a directly executable RoPE-compatible carrier.
	Other work explores RoPE-frequency selection~\citep{elitekv} and
	data-dependent low-rank compressibility~\citep{kvcore}, while
	KIVI~\citep{kivi}, KVQuant~\citep{kvquant}, Atom~\citep{atom},
	and QServe~\citep{qserve} primarily reduce numerical precision.

	\section{Conclusion}
	\label{sec:conclusion}
	
	GaugeRoPE characterizes and realizes a class of executable latent states under pretrained RoPE. \rev{The carrier preserves the pretrained rotary action, while an absorbed RKV correction supplies residual key and value information. Ablations show how compatible rank and complex coefficients improve attention fidelity and generation stability.} \rev{GaugeRoPE maintains quality close to the full-cache baseline at $5.85\times$ KV compression. Through direct latent execution, GaugeRoPE achieves up to $2.48\times$ the decode throughput of the vanilla uncompressed baseline.} \rev{These results establish direct executability as a design principle for post-training KV compression.}

	\bibliography{references}
	\bibliographystyle{references}
	
	\appendix
	\showrevisionsfalse

	\section{Proofs}
	\label{app:proofs}

	\paragraph{RoPE as a complex representation.}
	For each RoPE frequency $j$, one two-dimensional real rotation pair is
	identified with one complex coordinate. A real pair
	$(x_{2j-1},x_{2j})$ is written as $x_j=x_{2j-1}+ix_{2j}$, and position
	$p$ acts by multiplication with $e^{i\omega_jp}$. The real Euclidean
	inner product is
	\[
	\langle \mathbf{x},\mathbf{y}\rangle_{\mathbb{R}}
	=
	\sum_j \operatorname{Re}(\overline{x_j}y_j)
	=
	\operatorname{Re}(\mathbf{x}^\dagger \mathbf{y}).
	\]
	Thus the complex notation below is only a compact way to write real linear
	algebra on RoPE rotation pairs.
	
	\paragraph{Non-degeneracy assumption.}
	Let $\chi_j(p)=e^{i\omega_jp}$. We assume that the characters are pairwise
	distinct and that no retained character equals the conjugate of any retained
	character:
	\[
	\omega_j\not\equiv\omega_k\pmod {2\pi}\quad (j\neq k),
	\qquad
	\omega_j\not\equiv-\omega_k\pmod {2\pi}\quad (\text{all }j,k).
	\]
	The second condition also covers $j=k$ and therefore excludes
	self-conjugate frequencies ($2\omega_j\equiv0\pmod{2\pi}$). These are
	algebraic identities of position characters, not assumptions about a large
	numerical gap between nearby frequencies. If frequencies are exactly tied,
	the tied components merge into a larger multiplicity space.
	
	\subsection{Proof of Theorem~\ref{thm:intertwiner}}
	
	\begin{proof}
		\emph{Step 1: define the source and latent representations.}
		The stacked pre-RoPE key space is
		\[
		\mathcal{V}=\bigoplus_{j=1}^{F}\mathcal{V}_j,
		\qquad
		\mathcal{V}_j\cong\mathbb{C}^{D},
		\qquad
		D=L_gH_{kv}.
		\]
		Here $L_g$ denotes the number of repeated representation copies grouped in
		the abstract construction. The reported implementation uses $L_g=1$;
		cross-layer grouping is not part of the execution path evaluated in this
		paper. The dimension $D$ is the multiplicity of frequency $j$: each
		grouped layer/head slot contributes one complex coordinate at that
		frequency.
		RoPE acts on $\mathcal{V}_j$ by the scalar $e^{i\omega_jp}\mathbf{I}_D$. The latent space is
		\[
		\mathcal{V}'=\bigoplus_{j=1}^{F}\mathcal{V}'_j,
		\qquad
		\mathcal{V}'_j\cong\mathbb{C}^{r_j},
		\]
		and it carries the original frequency $j$ on $\mathcal{V}'_j$, namely
		$e^{i\omega_jp}\mathbf{I}_{r_j}$. A real-linear decoder $\mathbf{U}:\mathcal{V}'\to \mathcal{V}$ is
		RoPE-compatible when
		\[
		\mathbf{R}_p\mathbf{U}=\mathbf{U}\mathbf{R}'_p
		\qquad\text{for every position }p.
		\]
		
		\emph{Step 2: decompose the complex-linear part into frequency blocks.}
		Write $\mathbf{U}=(\mathbf{U}_{jk})_{j,k}$, where $\mathbf{U}_{jk}:\mathcal{V}'_k\to \mathcal{V}_j$. Taking the
		$(j,k)$ block of the intertwining equation gives
		\[
		e^{i\omega_jp}\mathbf{U}_{jk}
		=
		\mathbf{U}_{jk}e^{i\omega_kp}
		=
		e^{i\omega_kp}\mathbf{U}_{jk}.
		\]
		Therefore
		\[
		\bigl(e^{i\omega_jp}-e^{i\omega_kp}\bigr)\mathbf{U}_{jk}=0
		\qquad\text{for every }p.
		\]
		If $j\neq k$, non-degeneracy gives a position $p$ for which the scalar
		coefficient is nonzero, hence $\mathbf{U}_{jk}=0$. Thus an exact-compatible map
		cannot mix different RoPE frequencies.
		
		\emph{Step 3: rule out conjugate-frequency, antilinear couplings.}
		A general real linear map between complex coordinates can contain both a
		complex-linear part and an antilinear part:
		\[
		\mathbf{T}\mathbf{z}=\mathbf{A}\mathbf{z}+\mathbf{B}\overline{\mathbf{z}}.
		\]
		Intertwining the source frequency $k$ with the target frequency $j$
		requires
		\[
		e^{i\omega_jp}(\mathbf{A}\mathbf{z}+\mathbf{B}\overline{\mathbf{z}})
		=
		\mathbf{A}(e^{i\omega_kp}\mathbf{z})+\mathbf{B}(e^{-i\omega_kp}\overline{\mathbf{z}})
		\]
		for every position $p$. Matching the coefficients of $\mathbf{z}$ and
		$\overline{\mathbf{z}}$ gives
		\[
		(e^{i\omega_jp}-e^{i\omega_kp})\mathbf{A}=0,
		\qquad
		(e^{i\omega_jp}-e^{-i\omega_kp})\mathbf{B}=0.
		\]
		The first equation is the block condition above. The second forces $\mathbf{B}=0$
		unless $\omega_j=-\omega_k$ modulo $2\pi$. Under the stated
		non-degeneracy assumption, antilinear components vanish. Hence every
		compatible real map is represented by a complex-linear block matrix.
		
		\emph{Step 4: characterize the diagonal blocks.}
		For $j=k$, the block equation becomes
		\[
		e^{i\omega_jp}\mathbf{U}_{jj}=\mathbf{U}_{jj}e^{i\omega_jp},
		\]
		which is automatically true because both sides multiply by the same scalar.
		Therefore there is no additional constraint inside a same-frequency
		multiplicity space:
		\[
		\mathbf{U}_{jj}\in\mathbb{C}^{D\times r_j}
		\quad\text{is arbitrary.}
		\]
		Combining the off-diagonal exclusion with the diagonal freedom gives
		\begin{equation}
			\HomRoPE(\mathcal{V}',\mathcal{V})
			\cong
			\bigoplus_{j=1}^{F}\mathbb{C}^{D\times r_j}.
			\label{eq:hom-rope}
		\end{equation}
		
		\emph{Step 5: add the PCA/semi-unitary constraint.}
		The theorem characterizes all static linear intertwiners. If the decoder is
		additionally constrained to be a PCA-style semi-unitary basis, then each
		full-column-rank block satisfies
		\[
		\mathbf{U}_{jj}^{\dagger}\mathbf{U}_{jj}=\mathbf{I}_{r_j}.
		\]
		Thus the feasible block lies on the complex Stiefel manifold
		$\mathrm{St}_{\mathbb{C}}(r_j,D)$. This is an optimization constraint, not
		an additional RoPE constraint.
	\end{proof}
	
	\paragraph{Degenerate schedules.}
	If two frequencies are exactly tied, $\omega_j=\omega_k$, their multiplicity
	spaces should be combined before applying the theorem. If
	$\omega_j=-\omega_k$, conjugation can intertwine the two components; if
	$2\omega_j\equiv0\pmod{2\pi}$, an antilinear self-coupling is also
	possible. These cases enlarge the exact commutant rather than invalidate the
	intertwiner formulation.
	
	\subsection{Finite-window near-degeneracy and numerical conditioning}
	\label{app:finite-window}
	
	Theorem~\ref{thm:intertwiner} concerns exact equivariance for all integer
	positions. This is different from approximate equivariance on a bounded
	window $0\le p\le P$. Consider a complex-linear block
	$\mathbf{B}_{jk}:\mathcal{V}'_k\rightarrow \mathcal{V}_j$ between distinct frequencies. Its
	intertwining defect at position $p$ is
	\begin{equation}
		\|e^{i\omega_jp}\mathbf{B}_{jk}-\mathbf{B}_{jk}e^{i\omega_kp}\|
		=
		2\left|\sin\frac{(\omega_j-\omega_k)p}{2}\right|
		\|\mathbf{B}_{jk}\|.
		\label{eq:finite-window-linear-defect}
	\end{equation}
	Hence $P|\omega_j-\omega_k|\ll1$ can make a forbidden cross-frequency
	block nearly equivariant over that finite window even though it is not an
	exact intertwiner. For an antilinear block
	$\mathbf{C}_{jk}\overline{\mathbf{z}}$, the corresponding defect is
	\begin{equation}
		\|e^{i\omega_jp}\mathbf{C}_{jk}-\mathbf{C}_{jk}e^{-i\omega_kp}\|
		=
		2\left|\sin\frac{(\omega_j+\omega_k)p}{2}\right|
		\|\mathbf{C}_{jk}\|.
		\label{eq:finite-window-antilinear-defect}
	\end{equation}
	Thus conjugate or low-frequency characters may also support approximate
	antilinear couplings on a finite context. GaugeRoPE uses the exact commutant,
	which yields a positional rule independent of the evaluation horizon.
	
	Near-degenerate characters also do not create an ill-conditioned frequency
	separation step in GaugeRoPE. The model's RoPE indices determine the
	frequency groups exactly, after which each carrier is estimated from its
	own same-frequency Gram matrix $\mathbf{G}_j$. No linear system involving
	$(\omega_j-\omega_k)^{-1}$ or $(\omega_j+\omega_k)^{-1}$ is solved.
	Numerical sensitivity of the learned carrier is therefore governed by the
	eigenspectrum of $\mathbf{G}_j$, not by the spacing between different RoPE
	frequencies. If leading eigenvalues of $\mathbf{G}_j$ are repeated or nearly
	repeated, individual eigenvectors need not be identifiable, but any basis of
	the corresponding retained eigenspace remains inside the same exact-compatible
	frequency block.
	
	\subsection{Proof of Proposition~\ref{prop:gauge}}
	
	\begin{proof}
		\emph{Step 1: polar factorization of a complex rank-one carrier.}
		Fix a frequency $j$ and a unit carrier
		$\mathbf{u}_j\in\mathbb{C}^{H_{kv}}$. For each KV head $h$, write
		\[
		u_{h,j}=e^{i\phi_{h,j}}|u_{h,j}|,
		\]
		using any phase when $u_{h,j}=0$. Define
		\[
		\mathbf{D}_j=\operatorname{diag}(e^{i\phi_{1,j}},\ldots,e^{i\phi_{H_{kv},j}}),
		\qquad
		\mathbf{a}_j=(|u_{1,j}|,\ldots,|u_{H_{kv},j}|)^\top.
		\]
		Then $\mathbf{a}_j\in\mathbb{R}^{H_{kv}}_{\ge 0}$, $\|\mathbf{a}_j\|_2=1$, and
		$\mathbf{u}_j=\mathbf{D}_j\mathbf{a}_j$.
		
		\emph{Step 2: write the RoPE logit with the conjugate explicitly.}
		Let $q_{t,g,j}$ be the query coordinate of query head $g$ at time $t$,
		and let $\kappa_{s,h,j}$ be the pre-RoPE key coordinate of KV head $h$
		at time $s$. If query head $g$ attends to KV head $h=\pi(g)$, the
		frequency-$j$ contribution to the real attention logit is
		\begin{equation}
			\ell_{t,s,g,j}
			=
			\operatorname{Re}\!\left[
			\overline{q_{t,g,j}}\,
			e^{i\omega_j(s-t)}
			\kappa_{s,h,j}
			\right].
			\label{eq:app-complex-logit}
		\end{equation}
		
		\emph{Step 3: show gauge invariance of every logit term.}
		Apply the same phase change to the query and key coordinate:
		\[
		\widetilde q_{t,g,j}=e^{-i\phi_{h,j}}q_{t,g,j},
		\qquad
		\widetilde \kappa_{s,h,j}=e^{-i\phi_{h,j}}\kappa_{s,h,j},
		\qquad h=\pi(g).
		\]
		Substituting into Equation~(\ref{eq:app-complex-logit}),
		\[
		\operatorname{Re}\!\left[
		\overline{\widetilde q_{t,g,j}}\,
		e^{i\omega_j(s-t)}
		\widetilde \kappa_{s,h,j}
		\right]
		=
		\operatorname{Re}\!\left[
		\overline{q_{t,g,j}}\,
		e^{i\omega_j(s-t)}
		\kappa_{s,h,j}
		\right].
		\]
		The equality holds term by term, hence also after summing over frequencies
		and heads. In a GQA model, all query heads mapped to the same KV head receive
		that KV head's phase at the same frequency.
		
		\emph{Step 4: realize the gauge as an offline weight update.}
		Multiplication by $e^{-i\phi_{h,j}}$ is the real $2\times2$ rotation matrix
		of angle $-\phi_{h,j}$ on the corresponding RoPE plane. The gauge can
		therefore be folded into the learned Q and K projection weights before
		inference. It introduces no runtime complex arithmetic and does not change
		the pretrained model's attention function.
		
		\emph{Step 5: prove rank-one equivalence.}
		Compressing the original key vector with the complex carrier gives
		\[
		\mathbf{u}_j^\dagger \boldsymbol{\kappa}_j
		=
		(\mathbf{D}_j\mathbf{a}_j)^\dagger\boldsymbol{\kappa}_j
		=
		\mathbf{a}_j^\top \mathbf{D}_j^\dagger\boldsymbol{\kappa}_j
		=
		\mathbf{a}_j^\top \widetilde{\boldsymbol{\kappa}}_j.
		\]
		Thus the complex rank-one projection $\mathbf{u}_j$ in the original gauge is identical
		to the real rank-one projection $\mathbf{a}_j$ after phase alignment. Conversely, a
		real carrier is recovered by choosing $\mathbf{D}_j=\mathbf{I}$. Hence the phase-aligned real
		and complex rank-one families are exactly gauge-equivalent:
		\[
		\textit{phase-aligned real rank-1}
		\simeq_{\rm gauge}
		\textit{complex rank-1}.
		\]
		
		\emph{Step 6: strictness relative to the ungauged real family.}
		The ungauged real family requires the carrier to be real in the original
		coordinate system. A unit vector such as $\mathbf{u}=(1,i)^\top/\sqrt{2}$ has non-constant
		coordinate phases and cannot be represented by any real vector in that
		original gauge. Therefore the naive real rank-one family is a strict subset
		of the phase-aligned family.
	\end{proof}
	
	\subsection{Proof of Corollary~\ref{cor:absorb}}
	
	\begin{proof}
		\emph{Assumptions.}
		The corollary concerns static linear compression and exact execution relative
		to the compressed reconstruction. The maps are
		\[
		\mathbf{E}:\mathcal{V}\to \mathcal{Z},\qquad \mathbf{U}:\mathcal{Z}\to \mathcal{V},\qquad \mathbf{Q}:\mathcal{V}\to \mathcal{Z}.
		\]
		Here $\mathbf{E}$ is the encoder, $\mathbf{U}$ is the decoder, and $\mathbf{Q}$ is the query-side map
		used by absorbed attention. We assume:
		\begin{enumerate}
			\item $\mathbf{E}$ has full row rank, so $\mathcal{Z}=\operatorname{im}\mathbf{E}$ is the complete
			latent image.
			\item The decoder is restricted to this image; there are no unused latent
			directions outside $\operatorname{im}\mathbf{E}$.
			\item The latent space uses the original-frequency rotations $\mathbf{R}'_p$.
			\item Query-side equality is required for all $\mathbf{q},\mathbf{k},t,s$, not merely on a
			finite calibration set.
		\end{enumerate}
		The reconstructed key is $\hat{\mathbf{k}}_s=\mathbf{U}\mathbf{E}\mathbf{k}_s$.
		
		\emph{Sufficiency.}
		Assume first that $\mathbf{U}$ is an intertwiner:
		\[
		\mathbf{R}_p\mathbf{U}=\mathbf{U}\mathbf{R}'_p
		\qquad\text{for every }p.
		\]
		Write $\mathbf{z}_s=\mathbf{E}\mathbf{k}_s$ and define $\mathbf{c}_s=\mathbf{R}'_s\mathbf{z}_s$. Since the
		intertwining relation holds for the full rotation group, its adjoint form
		gives $\mathbf{U}^\dagger \mathbf{R}_t=\mathbf{R}'_t\mathbf{U}^\dagger$. For any query $\mathbf{q}$ at position $t$,
		\begin{align}
			\langle \mathbf{R}_t\mathbf{q},\mathbf{R}_s\hat{\mathbf{k}}_s\rangle_{\mathbb{R}}
			&=
			\langle \mathbf{R}_t\mathbf{q},\mathbf{R}_s\mathbf{U}\mathbf{z}_s\rangle_{\mathbb{R}} \\
			&=
			\langle \mathbf{R}_t\mathbf{q},\mathbf{U}\mathbf{R}'_s\mathbf{z}_s\rangle_{\mathbb{R}} \\
			&=
			\langle \mathbf{U}^\dagger \mathbf{R}_t\mathbf{q},\mathbf{R}'_s\mathbf{z}_s\rangle_{\mathbb{R}} \\
			&=
			\langle \mathbf{R}'_t\mathbf{U}^\dagger \mathbf{q},\mathbf{R}'_s\mathbf{z}_s\rangle_{\mathbb{R}} \\
			&=
			\langle \mathbf{R}'_t\mathbf{U}^\dagger \mathbf{q},\mathbf{c}_s\rangle_{\mathbb{R}}.
		\end{align}
		The implementation caches the rotated carrier $\mathbf{c}_s$.
		Equivalently, the score can be written using unrotated latent coordinates:
		\[
		\langle \mathbf{R}'_t\mathbf{U}^\dagger \mathbf{q},\mathbf{R}'_s\mathbf{z}_s\rangle_{\mathbb{R}}
		=
		\langle \mathbf{U}^\dagger \mathbf{q},\mathbf{R}'_{s-t}\mathbf{z}_s\rangle_{\mathbb{R}}.
		\]
		Thus $\mathbf{Q}=\mathbf{U}^\dagger$ provides exact absorbed scoring for any encoder $\mathbf{E}$
		with latent image $\mathcal{Z}$; the semi-unitary case used by Gauge has
		$\mathbf{E}=\mathbf{U}^\dagger$. Neither form materializes the dense historical key
		$\mathbf{R}_s\hat{\mathbf{k}}_s$.
		
		The value branch is compatible with the same execution pattern. Once the
		attention weights are computed from latent scores, values are aggregated
		linearly in latent space and decoded once after the weighted sum.
		
		\emph{Necessity.}
		Now suppose static maps $(\mathbf{E},\mathbf{U},\mathbf{Q})$ execute exactly against compressed
		reconstructions:
		\begin{equation}
			\langle \mathbf{R}_t\mathbf{q},\mathbf{R}_s\mathbf{U}\mathbf{E}\mathbf{k}\rangle_{\mathbb{R}}
			=
			\langle \mathbf{R}'_t\mathbf{Q}\mathbf{q},\mathbf{R}'_s\mathbf{E}\mathbf{k}\rangle_{\mathbb{R}}
			\label{eq:app-exact-execution}
		\end{equation}
		for every $\mathbf{q},\mathbf{k},t,s$. Set $t=s=0$. Then
		\[
		\langle \mathbf{q},\mathbf{U}\mathbf{E}\mathbf{k}\rangle_{\mathbb{R}}
		=
		\langle \mathbf{Q}\mathbf{q},\mathbf{E}\mathbf{k}\rangle_{\mathbb{R}}.
		\]
		Because $\mathbf{E}$ is onto $\mathcal{Z}$, every $\mathbf{z}\in \mathcal{Z}$ can be written as $\mathbf{z}=\mathbf{E}\mathbf{k}$ for some
		$\mathbf{k}$. Therefore
		\[
		\langle \mathbf{q},\mathbf{U}\mathbf{z}\rangle_{\mathbb{R}}
		=
		\langle \mathbf{Q}\mathbf{q},\mathbf{z}\rangle_{\mathbb{R}}
		\qquad\text{for all }\mathbf{q},\mathbf{z},
		\]
		which forces $\mathbf{Q}=\mathbf{U}^\dagger$.
		
		Substitute $\mathbf{Q}=\mathbf{U}^\dagger$ into Equation~(\ref{eq:app-exact-execution}). Using
		orthogonality of $\mathbf{R}_p$ and $\mathbf{R}'_p$, write the equality for relative offset
		$r=s-t$:
		\[
		\langle \mathbf{q},\mathbf{R}_r\mathbf{U}\mathbf{E}\mathbf{k}\rangle_{\mathbb{R}}
		=
		\langle \mathbf{q},\mathbf{U}\mathbf{R}'_r\mathbf{E}\mathbf{k}\rangle_{\mathbb{R}}
		\qquad\text{for all }\mathbf{q},\mathbf{k},r.
		\]
		Since this holds for all $\mathbf{q}$,
		\[
		(\mathbf{R}_r\mathbf{U}-\mathbf{U}\mathbf{R}'_r)\mathbf{E}\mathbf{k}=0
		\qquad\text{for all }\mathbf{k},r.
		\]
		Using again that $\mathbf{E}$ is onto $\mathcal{Z}$, we obtain
		\[
		\mathbf{R}_r\mathbf{U}=\mathbf{U}\mathbf{R}'_r
		\qquad\text{on all of }\mathcal{Z},\text{ for every }r.
		\]
		This is exactly the intertwining condition.
		
	\end{proof}

	\subsection{GQA query realization and higher-rank realification}
	\label{app:gqa-realization}
	
	For query head $g$, let $h=\pi(g)$ be its shared KV head and let
	$\mathbf{e}_h\in\mathbb{C}^{D}$ select that coordinate in a same-frequency block. With
	$\mathbf{z}_{s,j}=\mathbf{U}_j^\dagger\boldsymbol{\kappa}_{s,j}$, define
	\begin{equation}
		\widetilde{\mathbf{q}}_{t,g,j}=\mathbf{U}_j^\dagger \mathbf{e}_h\,q_{t,g,j}.
	\end{equation}
	Then the carrier reconstruction obeys
	\begin{equation}
		\operatorname{Re}\!\left[
		\overline{q_{t,g,j}}e^{i\omega_j(s-t)}\mathbf{e}_h^\top \mathbf{U}_j \mathbf{z}_{s,j}
		\right]
		=
		\operatorname{Re}\!\left[
		\widetilde{\mathbf{q}}_{t,g,j}^\dagger e^{i\omega_j(s-t)}\mathbf{z}_{s,j}
		\right],
	\end{equation}
	which is the GQA form of absorbed carrier scoring.
	
	Proposition~\ref{prop:gauge} gives the stronger phase-gauged real-scalar
	realization only at rank one. For $r_j>1$, including the reported R2 carrier,
	we use the standard realification of a complex-linear block. Each complex
	coefficient $a+ib$ acts through
	\begin{equation}
		a+ib
		\quad\longleftrightarrow\quad
		\begin{bmatrix}a&-b\\ b&a\end{bmatrix}.
		\label{eq:complex-realification}
	\end{equation}
	The cached complex coefficients are stored as their real and imaginary parts,
	matching the ordinary real-scalar cache representation used by the
	implementation.
	
	\section{Calibration and Construction Details}
	\label{app:calibration}
	
GaugeRoPE uses unlabeled WikiText-103 raw training data (\texttt{calib.txt}), excluding evaluation data; Appendix~\ref{app:sample-sensitivity} also tests WikiText-2 calibration. Calibration uses approximately 200K carrier tokens and 100K joint-RKV tokens in 2048-token blocks, counted with each backbone's tokenizer; model weights remain frozen. Carrier bases are fitted independently for each layer from the same-frequency Gram matrices in Equation~(\ref{eq:gram}).
	
	\paragraph{Residual/value moments and energy balancing.}
	Using the residual in Equation~(\ref{eq:key-residual}), let $m=H_{kv}d_h$ and $\mathbf{x}(t)=[\mathbf{r}^K(t);\mathbf{v}(t)]\in\mathbb{R}^{2m}$. From the second-moment matrix $\mathbf{C}_0=\mathbb{E}[\mathbf{x}\mathbf{x}^\top]$, define the per-coordinate branch
	energies
	\begin{equation}
		e_k=\frac{\operatorname{tr}((\mathbf{C}_0)_{:m,:m})}{m},
		\qquad
		e_v=\frac{\operatorname{tr}((\mathbf{C}_0)_{m:,m:})}{m},
	\end{equation}
	and
	\begin{equation}
		\mathbf{S}=\operatorname{diag}\!\left(e_k^{-1/2}\mathbf{I}_m,\;e_v^{-1/2}\mathbf{I}_m\right).
		\label{eq:rkv-balancing}
	\end{equation}
	We eigendecompose $\widetilde{\mathbf{C}}=\mathbf{S}\mathbf{C}_0\mathbf{S}$ and retain its leading
	$\mathbf{W}_c\in\mathbb{R}^{2m\times d_c}$. The cached code and reconstruction are
	\begin{equation}
		\mathbf{b}=\mathbf{W}_c^\top \mathbf{S}\mathbf{x},
		\qquad
		\hat{\mathbf{x}}=\mathbf{S}^{-1}\mathbf{W}_c\mathbf{b}.
		\label{eq:rkv-pca-codec}
	\end{equation}
	Thus $\mathbf{A}_R=\mathbf{S}\mathbf{W}_c$ in Equation~(\ref{eq:rkv-codec}), while the decoder is
	$\mathbf{D}_R=\mathbf{S}^{-1}\mathbf{W}_c=[\mathbf{M}_K^\top\;\mathbf{M}_V^\top]^\top$.

\section{Additional Quality Controls}
\label{app:quality-robustness}
\subsection{Recent-Token Precision}
\label{app:window-ablation}
Table~\ref{tab:window-summary} compares native recent-token budgets. At length $T$, cache bytes per token average $B+(S-B)\min(W,T)/T$, for compressed/native sizes $B/S$ and window $W$; shared scales are counted separately. Unmarked main-table GaugeRoPE configurations and quantized decode compress all history. Cross-backbone, NIAH length-scaling, and BF16 system evaluations retain 256 native recent tokens, including the matched TransMLA comparison.

\begin{table}[h]\centering\small
	\caption{Recent-token ablation. BF16 uses the original 3,750-example LongBench evaluation; 8/4 bits uses the main-table samples and mean measured compression, including native tokens and metadata.}
	\label{tab:window-summary}
	\begin{tabular*}{\linewidth}{@{\extracolsep{\fill}}lrrrr@{}}\toprule
		Cached scalars & Native recent tokens & Comp. Ratio & RULER & LongBench \\\midrule
		1024 & 0 & $2.00\times$ & 85.92 & 36.46 \\
		1024 & 256 & $1.94\times$ & 86.69 & 38.84 \\
        \midrule
        \multicolumn{5}{@{}l}{\textbf{Gauge-50\% (8/4 bits)}} \\
        Native recent tokens & \multicolumn{2}{c}{Comp. Ratio (RULER / LB)} & RULER & LongBench \\
        0 & \multicolumn{2}{c}{$5.85\times$ / $5.85\times$} & 85.56 & 35.73 \\
        128 & \multicolumn{2}{c}{$5.42\times$ / $5.41\times$} & 86.31 & 38.63 \\
        256 & \multicolumn{2}{c}{$5.04\times$ / $5.07\times$} & 86.41 & 38.92 \\
		\bottomrule
	\end{tabular*}
\end{table}

Mixed precision retains exactly $W$ finalized tokens; the earlier BF16 control evicts whole blocks. At $W=256$, LongBench gains 3.19 points over no window (paired 95\% CI $[1.79,4.63]$), only 0.28 above $W=128$ (CI $[-0.65,1.22]$), mainly through code completion.

\subsection{Unconstrained Carrier Control}
\label{app:generic-matched}
Compatibility preserves direct execution while improving retrieval over unconstrained PCA. At 1024 real scalars per token/layer, GaugeRoPE outperforms an unconstrained rank-256 carrier with refitted RKV by 7.73 RULER points (paired 95\% CI $[4.88,10.55]$). The unconstrained carrier scores slightly higher on the original 3,750-example LongBench evaluation (37.80 versus 36.46), but requires historical-key reconstruction for RoPE scoring.

\subsection{Quantization Details}
\label{app:quantization}
The mixed-precision configuration uses static symmetric 8-bit quantization for the carrier, with scales shared across each real/imaginary pair, and dynamic asymmetric 4-bit quantization for RKV in groups of 64 coordinates per token. Each RKV group stores an FP32 scale and a uint8 zero point. Packed codes and dynamic metadata require 700 B per layer/token, with an additional 32,768 B of shared carrier scales across the model. Without native recent tokens, including this metadata gives $5.843$--$5.851\times$ compression across the evaluated sequences. Palu and ReCalKV use their official Hadamard rotation and dynamic asymmetric 6-bit latent quantizers. Their quality results use fake quantization; compression ratios account for packed codes and an FP32 scale plus a uint8 zero point per low-rank group, giving 788 and 781.5 B per layer/token, respectively.

\section{Compatible Capacity and Hybrid Mechanism}
\label{app:capacity-mechanism}

\subsection{Rank and Operator Fidelity}
\label{app:r1-r2-operator}
Carrier-only comparisons isolate the added compatible mode; hybrid comparisons also refit RKV. The latter reduce KL by 0.03344 (95\% CI $[0.03303,0.03385]$). Diagnostics evaluate FP32 operator algebra on normalized Full states from 32 prompts, using the last 512 queries, all layers and heads, historical positions $s<t$, and native scaling $1/\sqrt{128}$.

For Full and tested attention distributions $p,\hat p$, we compute $D_{\mathrm{KL}}(p\|\hat p)=\sum_{s<t}p_s(\log p_s-\log\hat p_s)$ with FP32 log-softmax and no probability floor. KL and Top-8 overlap, $|\mathrm{Top8}(p)\cap\mathrm{Top8}(\hat p)|/8$, are averaged over queries and heads, then equally over layers and prompts. Output relative L2 uses Full values to isolate scoring error: each head's error norm over queries and output coordinates is divided by its Full-output norm, then averaged over heads, layers, and prompts.

Paired 95\% percentile intervals resample prompt pairs within each task and average tasks equally. Each prompt's layers, heads, and queries remain together; calibration intervals are computed separately for each fitted codec pair.

\subsection{Real versus Complex Coefficients}
\label{app:complex-control}
Real R2 uses the leading two eigenvectors of $\operatorname{Re}(\mathbf G_j)$ on both coordinate components, realizing the real-coefficient RoRoPE construction at rank two; Complex R2 uses those of the Hermitian $\mathbf G_j$. Both use the same calibration moments and native frequencies, with no frequency folding, and allocate 256 carrier and 768 RKV dimensions. Each carrier's residual and values are fitted by the same balanced joint PCA in Appendix~\ref{app:calibration}. Complex coefficients improve capture in all 2048 layer-frequency blocks and reduce uncaptured Gram energy by 18.3\%. LongBench-6 comprises PassageRetrieval-en, HotpotQA, Qasper, NarrativeQA, QMSum, and RepoBench-P, with 30 examples each.

\begin{table}[htbp]
\centering\small
\caption{Equal-rank real/complex control at the same cache size.}
\label{tab:complex-control}
\begin{tabular*}{.94\linewidth}{@{\extracolsep{\fill}}lrr@{}}
\toprule
Metric & Real R2 & Complex R2 \\
\midrule
Spectral capture (\%) & 79.27 & 83.06 \\
Carrier-only KL & 0.472319 & 0.406088 \\
Hybrid KL & 0.080179 & 0.072916 \\
Hybrid Top-8 & 0.813745 & 0.825609 \\
Hybrid output rel-L2 & 0.267827 & 0.251463 \\
RULER (13 tasks) & 6.05 & 85.92 \\
LongBench-6 & 0.13 & 42.98 \\
\bottomrule
\end{tabular*}
\end{table}

Execution checks on two prompts find exact cache encoding and dense/direct agreement within $3.4\times10^{-6}$ relative L2 for scores and value aggregation. Real R2 reconstructs values slightly more accurately on the shared initial states, yet its trajectory subsequently diverges during prefill. First-layer hybrid KL is approximately 0.170 for Real R2 and 0.074 for Complex R2; Complex improves KL in 25 of 32 layers for each calibration sample.

\subsection{Residual Positional Error}
\label{app:residual-attribution}
Restoring residual RoPE improves attention fidelity in every prompt and layer mean (Table~\ref{tab:residual-attribution}). With compressed codes and Full values fixed, the remaining gap to the exact reference isolates residual compression error. The positional discrepancy grows with distance: across four bins from $(0,256)$ to $[4\mathrm{K},8\mathrm{K})$, A--B scaled-logit MAE rises from 0.3492 to 0.5710, while B--C stays near 0.16. On three RULER tasks with 50 paired examples each, restoring residual RoPE changes the macro score by +3.07 points (95\% CI $[-1.07,7.27]$); this attribution path reconstructs residual keys.

\begin{table}[!htbp]
\centering\small
\caption{\textbf{Residual-RoPE attribution.} Fixed codes and Full values; the protocol matches Appendix~\ref{app:r1-r2-operator}.}
\label{tab:residual-attribution}
\begin{tabular*}{\linewidth}{@{\extracolsep{\fill}}lrrr@{}}
\toprule
Residual score & Attention KL $\downarrow$ & Top-8 $\uparrow$ & Output rel-L2 $\downarrow$ \\
\midrule
A: position independent & 0.072916 & 0.825609 & 0.251463 \\
B: RoPE restored & 0.025022 & 0.912959 & 0.153077 \\
C: exact reference & $1.55\times10^{-9}$ & 0.999999 & $9.39\times10^{-7}$ \\
\bottomrule
\end{tabular*}
\end{table}

\FloatBarrier
\subsection{Calibration-Sample and Corpus Sensitivity}
\label{app:sample-sensitivity}
We repeat calibration on three draws of 98 out of 196 source blocks (2048 tokens each; seeds 53--55). Each comparison shares the draw, token budgets, and balanced RKV fitting rule. Evaluation uses 20 examples each from \texttt{niah\_multikey\_1}, \texttt{niah\_multikey\_3}, \texttt{cwe}, and \texttt{qa\_1}. \rev{Real and Complex R2 have matched rank and storage; their operator diagnostics use two prompts per task, the last 128 queries, and all layers and heads on normalized Full states.}

\begin{table}[htbp]
\centering\small
\caption{Calibration-sample sensitivity at 1024 real scalars per token/layer. Four-task RULER scores; 95\% CIs for Complex R2 minus either control use 20,000 within-task paired bootstrap resamples.}
\label{tab:calibration-sensitivity}
\begin{tabular*}{\linewidth}{@{\extracolsep{\fill}}lrrrl@{}}
\toprule
Seed & R1 & Real R2 & Complex R2 & 95\% CI for gain \\
\midrule
53 & 0.00 & 0.00 & 78.00 & $[69.75,85.75]$ \\
54 & 0.00 & 0.00 & 84.75 & $[77.62,91.25]$ \\
55 & 0.00 & 0.00 & 80.00 & $[72.25,87.12]$ \\
\bottomrule
\end{tabular*}
\end{table}

The closed-loop advantage in Table~\ref{tab:calibration-sensitivity} is accompanied by consistent spectral and operator gains. Across all calibration draws, R2 reduces carrier-only and hybrid KL relative to R1 by approximately 39\% and 31\%. At fixed rank two, complex coefficients add approximately 3.8 percentage points of capture and improve KL, Top-8 overlap, and output error for every diagnostic prompt.

The advantage also persists with WikiText-2, Palu's calibration corpus. Under the same fitting procedure and token budgets, R2/R1 score 82.13/0.00 (paired gain 95\% CI $[74.88,89.00]$), with carrier-only and hybrid KL reductions of 39.05\% and 31.43\%.

\showrevisionsfalse
	\section{MHA Generality at Fixed State Budget}
\label{app:mha-generality}
Increasing the carrier's share of a fixed cache budget improves CodeLlama (Table~\ref{tab:mha-generality-category}). It uses 4096 latent scalars per token/layer, assigning fraction $\alpha$ to the carrier and the remainder to RKV.

\begin{table}[htbp]\centering\small
\caption{RULER macros for the fixed-budget MHA rank sweep.}
\label{tab:mha-generality-category}
\begin{tabular*}{\linewidth}{@{\extracolsep{\fill}}lrrrr@{}}\toprule
Backbone & Full & $\alpha=1/16$ & $\alpha=1/8$ & $\alpha=1/4$ \\\midrule
CodeLlama & 79.57 & 58.80 & 67.78 & 69.50 \\
\bottomrule\end{tabular*}\end{table}

\section{Baseline Configurations}
	\label{app:baseline-details}
	
	Palu and ReCalKV quality runs use WikiText-2 calibration and match GaugeRoPE-50\%'s $2\times$ budget, evaluation prompts, generation settings, and scorers. The TransMLA configuration is given in Appendix~\ref{app:additional-quality}.

\label{app:palu-repro}\label{app:recalkv-repro}
Palu uses its released whitening and low-rank decomposition procedure~\citep{palu}. ReCalKV follows its official HSR, value calibration, and fusion procedure~\citep{recalkv}, with Fisher-uniform rank allocation and head-group size 4. Calibration and updating each use 256 WikiText-2 samples, of length 1024 and 2048 respectively; learned head permutations are preserved during evaluation.

\FloatBarrier
\section{System Implementation and Measurement}
\label{app:system-protocol}
Measurements use H100 PCIe, BF16 computation for Full and the low-rank methods. Full uses PyTorch FlashAttention. Batch-4 full-model throughput reports medians of six 128-step requests after one warmup on matched inputs. Timing includes forward, argmax, and cache updates, excluding prefill, without CUDA graphs or \texttt{torch.compile}; BF16 throughput ranges are below 1\%. Attention latency includes projections and output mapping, averaging per-layer medians over three rounds of 128 calls after one warmup. Peak memory includes weights, codec, cache, and temporary tensors.

Gauge combines FP32 carrier and RKV scores before split-PV, which fuses exponential weighting and latent aggregation without materializing attention weights. Palu and ReCalKV use fused reconstruction kernels. These are the execution paths used for the reported BF16 measurements.

\begin{table}[htbp]
\centering\small
\caption{\textbf{Larger batches and models.} Full/Gauge medians over three requests after one warmup.}
\label{tab:scalability}
\begin{tabular*}{\linewidth}{@{\extracolsep{\fill}}lrrr@{}}
\toprule
Model / batch / context & Throughput (tok/s) & Speedup & Decode peak (GiB) \\
\midrule
Mistral-7B / 8 / 16K & 39.69 / 91.77 & $2.31\times$ & 31.67 / 24.42 \\
Mistral-7B / 16 / 16K & 41.76 / 105.00 & $2.51\times$ & 49.82 / 34.18 \\
Qwen2.5-14B-Instruct / 4 / 32K & 11.91 / 26.92 & $2.26\times$ & 54.29 / 43.21 \\
\bottomrule
\end{tabular*}
\end{table}

At batch size one, Mistral Gauge/Full throughput is $0.56\times$ at 8K and $1.29\times$ at 32K; longer histories and larger batches amortize fixed projection and launch costs.

\section{Additional Quality Results}
\label{app:additional-quality}
\begin{table}[htbp]
\centering\small
\caption{\textbf{NIAH length scaling.} Four tasks, 20 examples each. Latent budgets: 1024 real scalars per token/layer for Mistral-7B and 512 for Qwen2.5-7B.}
\label{tab:niah-length}
\begin{tabular*}{\linewidth}{@{\extracolsep{\fill}}llrrr@{}}
\toprule
Model & Method & 8K & 16K & 32K \\
\midrule
\multirow{2}{*}{Mistral} & Full & 92.50 & 88.75 & 90.63 \\
 & Gauge & \textbf{86.56} & \textbf{82.19} & \textbf{80.31} \\
\midrule
\multirow{3}{*}{Qwen} & Full & 96.88 & 95.94 & 93.13 \\
 & TransMLA & 76.25 & 37.19 & 11.88 \\
 & Gauge & \textbf{90.00} & \textbf{83.75} & \textbf{78.75} \\
\bottomrule
\end{tabular*}

\end{table}

Table~\ref{tab:niah-length} evaluates \texttt{niah\_single\_2}, \texttt{niah\_multikey\_1}, \texttt{niah\_multiquery}, and \texttt{niah\_multivalue}. TransMLA uses FreqFold=4 with 128 positional and 384 joint-BKV scalars per token/layer, matching Gauge's budget and inputs without uptraining. On Qwen at 32K, increasing the carrier/RKV allocation from 128/128 to 256/128 and then 256/256 raises NIAH scores from 25.31 to 58.75 and 78.75, showing the complementary benefits of carrier and residual capacity.

\end{document}

%% file: math_commands.tex
\usepackage{amsmath,amsfonts,bm}

\def\eqref#1{equation~\ref{#1}}

\def\1{\bm{1}}

\DeclareMathAlphabet{\mathsfit}{\encodingdefault}{\sfdefault}{m}{sl}
\SetMathAlphabet{\mathsfit}{bold}{\encodingdefault}{\sfdefault}{bx}{n}

